\documentclass{article}
\usepackage[english]{babel}
\usepackage{url}
\usepackage[letterpaper,top=2cm,bottom=2cm,left=3cm,right=3cm,marginparwidth=1.75cm]{geometry}

\usepackage{amsmath}
\usepackage{graphicx}
\usepackage[colorlinks=true, allcolors=blue]{hyperref}

\title{US \& MRI Image Fusion Based on Markerless Skin Registration}
\author{Martina Paccini, Giacomo Paschina, Stefano De Beni, Andrei Stefanov,\\
	 Velizar Kolev, Giuseppe Patanè}

\begin{document}
	\maketitle
	
	\begin{abstract}
		This paper presents an innovative automatic fusion imaging system that combines 3D CT/MR images with real-time ultrasound (US) acquisition. The system eliminates the need for external physical markers and complex training, making image fusion feasible for physicians with different experience levels.
		The integrated system involves a portable 3D camera for patient-specific surface acquisition, an electromagnetic tracking system, and US components. The fusion algorithm comprises two main parts: skin segmentation and rigid co-registration, both integrated into the US machine. The co-registration software aligns the surface extracted from CT/MR images with patient-specific coordinates, facilitating rapid and effective fusion.
		Experimental testing in different settings validates the system's accuracy, computational efficiency, noise robustness, and operator independence. The co-registration error remains under the acceptable range of~$1$ cm.		
	\end{abstract}
	%
	%\tableofcontents
	%
	\section{Introduction}
	\label{sec:introduction}
	Medical imaging offers many image acquisition techniques, which allow us to obtain information related to different tissues with various settings, such as signal-to-noise ratio, contrast, and resolution. Generally, high-resolution imaging has the drawback of requiring extended image acquisition time, thus making these techniques unsuitable for real-time image processing and analysis. For instance, surgical tools guidance, requires monitoring and guiding the insertion of a biopsy needle. Similarly, in cardiological imaging, ongoing monitoring of organ functional reactions is crucial for evaluating heart function. In contrast to high-resolution imaging (e.g., CT, PET, MRI), US imaging allows real-time acquisition. It assists physicians in various interventional applications, from simple biopsy to more thorough procedures like mini-invasive tumour treatment or neurosurgery. However, the US has a reduced field of view compared to other imaging techniques and a lower quality of features, such as image resolution or the revelation and reproduction of certain kinds of tissue, e.g., soft tissues. Therefore, medical imaging is moving toward combining real-time US with other acquisition modalities. This image combination, or fusion, is applied to several applications, like diagnostics and mini-invasive surgical interventions.
	
	The core concept in fusion imaging is accurately registering the patient's anatomy in different medical imaging data, such as US, CT, or MR images, which implies aligning different acquisitions into a standard reference system. This process often leads to overcomplicated systems for an actual clinical application. Image fusion (Sect.~\ref{sec:RELATED-WORK}) is carried out by tracking the US probe's position, orientation, and displacements during the acquisition within a reference system common to the US images and the other modalities considered. This setting implies the use of probe trackers of different natures: probe trackers based on (i) optical technology can simultaneously track several objects with high precision but with the drawback of the line of sight that is difficult to guarantee in an interventional room. Alternatively, (ii) EM technology, whose drawback is the high sensitivity to metals plus the necessity of several wires, one for each object that needs to be tracked. Moreover, typically EM tracking systems require choosing markers, on the patient with a wand (or needle guide wire or catheter) and simultaneously selecting those markers on the preprocedural image (CT/MRI) or leveraging fiducial patches on the procedural (CT/MRI) image ~\cite{abi2012multimodality}.
	
\begin{figure}[t]
		\centering
		\begin{tabular}{cc}
			(a)\includegraphics[width=0.45 \linewidth]{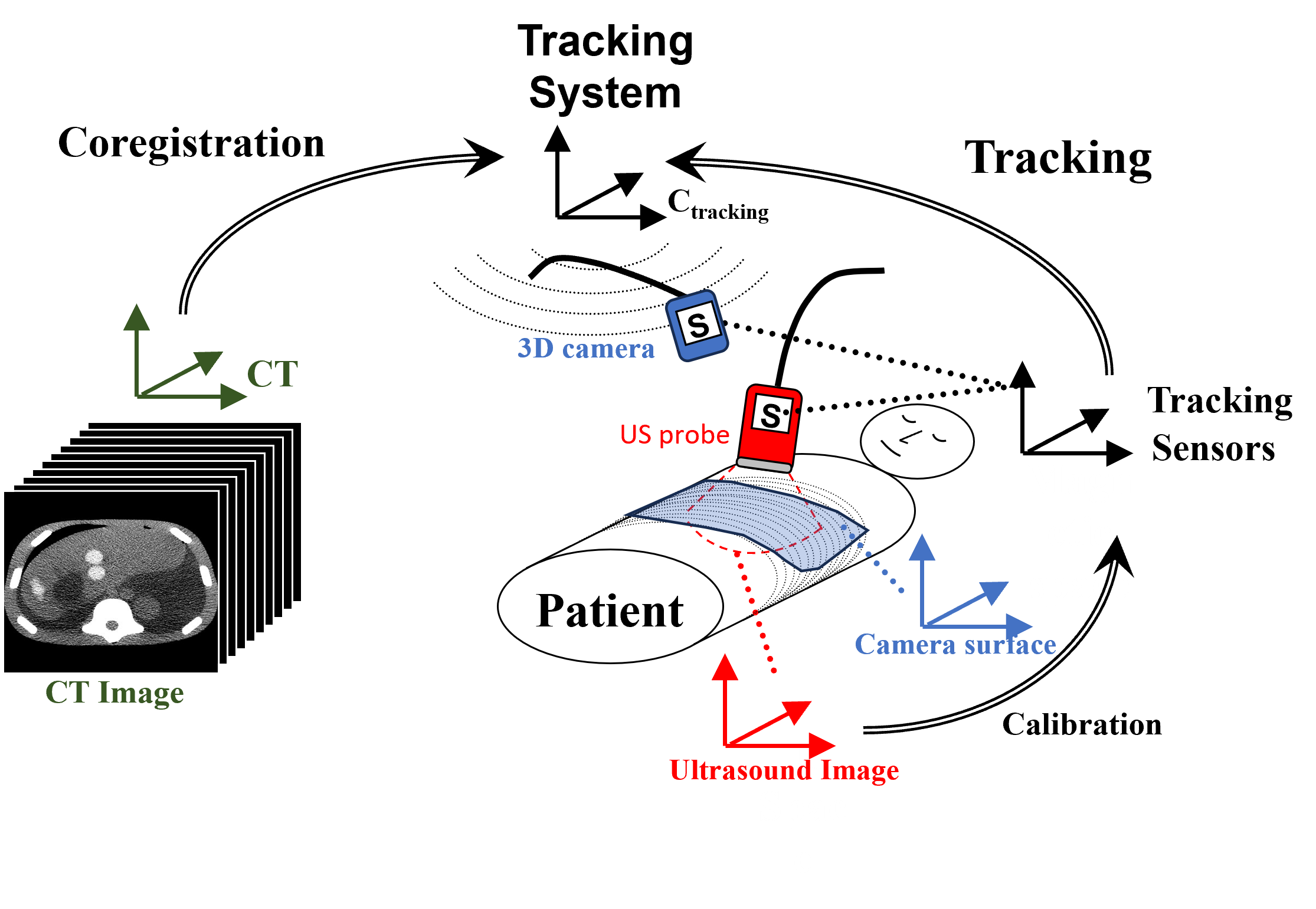}
			&(b)\includegraphics[width=0.4 \linewidth] {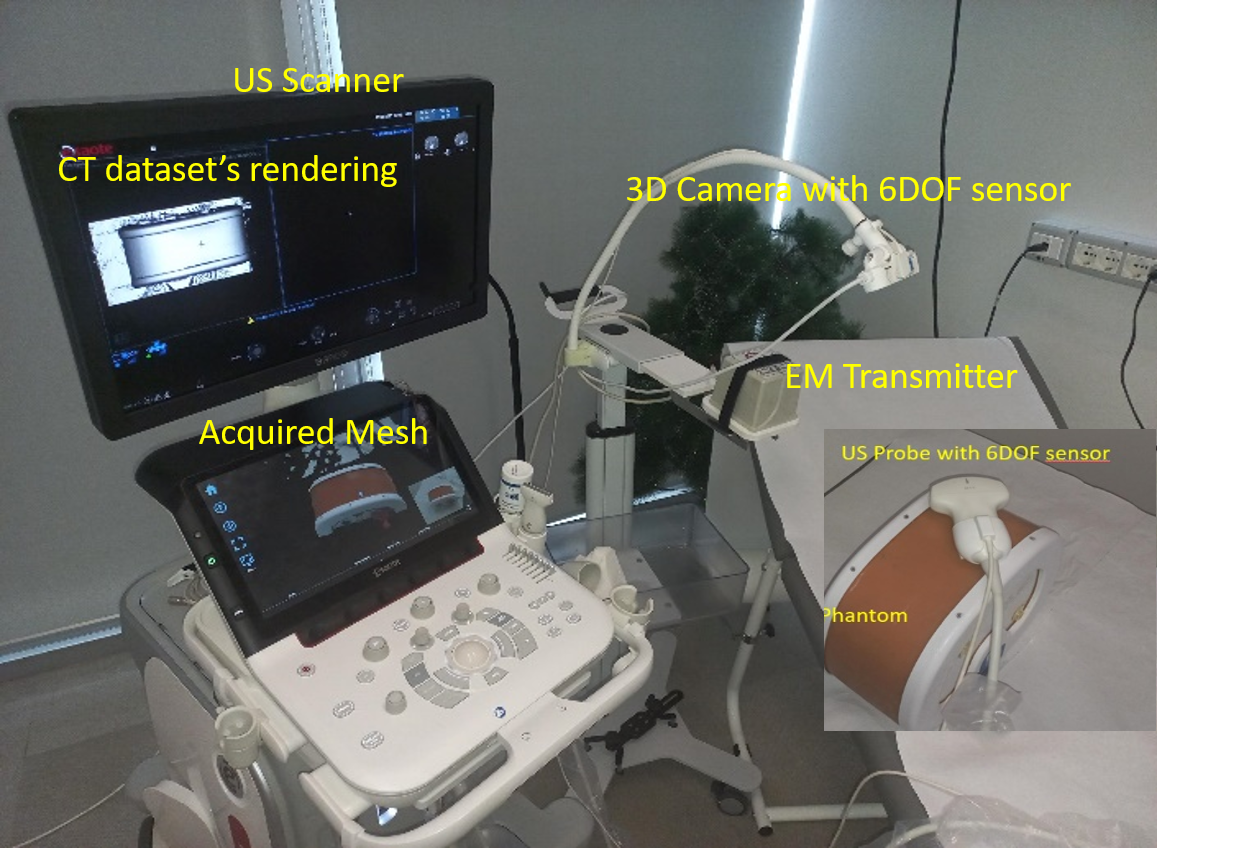}
		\end{tabular}
		\caption{\label{fig:hw_sw} (a) Hardware and software components of the system and their mutual interaction for the image fusion. (b) US system integrated into a testing setup that mimics the clinical environment.}
	\end{figure}

	This paper introduces an innovative \textit{image fusion system} (Fig.~\ref{fig:hw_sw}) that combines 3D CT/MR images with US acquisition in real-time (Sect.~\ref{sec:PROPOSED-APPROACH}). The system consists of hardware and software components designed to seamlessly integrate the resulting system into a clinical environment, particularly in interventional radiology, with a direct approach that does not require specific training. This efficient and intuitive integration is achieved trough the use of a 3D depth camera able to acquire a 3D surface of the subject undergoing the US exam as quickly as photograph-taking~\cite{DepthCam19:online}. Indeed, the surface obtained by the 3D camera bridges the 3D anatomy acquired by MR/CT and US images, allowing their fast fusion. The highly portable 3D camera can be introduced in an operating room without compromising the pre-existing set-up. The other hardware components of the image fusion system include a US system and a simplified EM tracking system that does not require the placement of physical markers or fiducial patches.
	
	The tracking system comprises an electronic unit, a mid-range transmitter and sensors for tracking traces of the position and coordinates of the 3D camera and the US probe. The transmitter generates an electromagnetic field and can simultaneously track up to four sensors at 70 times per second. The sensors are placed on the components to be tracked to establish their spatial relationship within the fusion imaging setup (Fig.~\ref{fig:hw_sw}). One sensor is connected to the US probe, and the other is associated with the 3D camera, allowing the representation of the US image and the 3D surface acquired by the camera in a unique coordinate system (i.e., the tracking coordinates).
	
	The software components can be divided into tracking, surface co-registration, and visualisation software. The system segments the skin surface from the CT/MRI and generates a 3D surface overlay on the patient's 3D rendering. Then, the clinician acquires the 3D skin with the 3D camera, and the \textit{co-registration software} enables the image fusion within a few seconds, together with the registration error visualisation, facilitating the identification of potential mismatches in the scan area. Upon successful registration, the system presents MRI or CT images alongside the US image in various visualisation modes. This allows the clinician to access the corresponding anatomical information and real-time US data during the examination.
	
	Differently from state-of-the-art methods, where the registration techniques are based mainly on the doctor's ability and require a long learning curve, the proposed image fusion is suitable even for radiologists with lower experience levels. Moreover, the proposed method avoids external physical markers, which, despite the interesting results~\cite{solbiati2018augmented}, could not comply with the existing hospital workflow since marker positioning requires time and can be error-prone. The skin segmentation method developed is highly general and can be applied to different anatomical regions and image types. Even the new AI-based methods for automatic liver shape or vessel tree segmentation require sequences that are only sometimes present in the patient data set and encounter even more complications due to the high variability of the images from series to series.
	
	The proposed image fusion system has been tested considering different aspects: the co-registration accuracy between MR/CT and US images (millimetric error), the computational cost for real-time applications (in seconds), noise robustness, and independence from the operator and setting (Sects.~\ref{sec:RESULTS},~\ref{sec:FUTURE-WORK}).
\section{Related work\label{sec:RELATED-WORK}}
\paragraph*{Fusion imaging}
Fusion imaging is in the guideline for several clinical procedures like targeted prostate biopsy that allow more comfort for the patient and more reliable results in the tissue sampling~\cite{bjurlin2016multiparametric,gayet2016value}. In abdominal applications, fusion imaging is widely used for liver tumour treatment using ablation techniques based on radiofrequency (RF) needles, microwave (MW) antennas, laser fibre or cryoprobes. All these techniques require the placement of ablation electrodes or applicators into the lesion and the deployment of energy until the tissue reaches a temperature over 65°C (RF MW Laser) or less than -18°C (Cryo), causing cellular death. The main challenge for fusion imaging is to reach a millimetric accuracy between the target in the MRI/CT and US real-time imaging due to different factors influencing the result. Firstly, physical phenomena can reduce the accuracy depending on the nature of the tracking system: an optical tracker can produce errors if the line of sight is not maintained, and EM ones can be influenced by metal distortion. Furthermore, the different postures of the patient between the CT/MR image acquisition and the current position during the percutaneous procedure under US guidance can negatively affect fusion accuracy. Even less predictable movement due to the patient's breathing and the different position of the organ between the CT/MR acquisition and the current examination has to be taken into account. Considering the liver, the patient's breathing generally induces a mismatch from the inspiration to the expiration phase, generating an error in targeting up to 5cm~\cite{lee2014fusion}.

Different clinical procedures aim to reduce the error sources by either driving the patient's breathing during the second modality examination or controlling (manually or aided by a breathing machine) the breathing movements during the procedure. The fusion imaging systems generally calibrate the organ directly to minimise possible errors. This solution fits well for those applications (neuro-oncology, muscle-skeletal) where the organ is not subjected to a deformation. The solutions introduced for deforming organs apply the registration based on elastic fusion, trying to deform the original image following the organ's shape or vessel tree depicted by a US image or a volume. Unfortunately, all these solutions are palliative since the deformation is applied without knowing the organ rigidity and assuming that rigidity is homogeneous inside the organ, where the nodule is generally harder than parenchyma, as the elastography image technique confirms. Additional sensors are also used to track the patient's breathing to synchronise the two image modalities in their best match during the breathing phase~\cite{li2019respiratory, santini2020ultrasound, madore2023external, yang2015subject}.
	
\paragraph*{Skin segmentation}
In breast image analysis, a few works segmented the skin as part of their pipeline. For the diagnosis of breast diseases with the dynamic contrast-enhanced MRI (DCE-MRI), the segmentation of the breast's first layer of skin has been obtained through a pre-processing of the image with median filters and mathematical morphology followed by the identification of the upper boundary of the breast, which is the skin boundary~\cite{lee2018automated}. However, the method used to identify the upper boundary is not explicitly described and focuses only on DCE-MRI images. Another technique for breast skin identification on classical CT and MRI can be obtained through thresholding followed by morphological filters~\cite{jermyn2013fast} or 3D vector-based connected component algorithm~\cite{wang2012fully}. Thresholds have also been leveraged in other districts on the raw image~\cite{baum2012does} or after a pre-processing aimed at edge enhancement~\cite{AnAut4958341:online}. These works apply the thresholding method on the whole image to classify the pixel in the background (black) or body (white) and then use other filtering methods to clean the obtained result. Skin segmentation is generally applied to CT images, as the corresponding skin's Hounsfield Unit (HU) value is known~\cite{baum2012does} and cannot be applied directly to other imaging modalities. In ~\cite{beare2016automated}, the Watershed transform from markers has been used to a gradient image containing light to do dark transitions obtained from T1-MRI. In deep learning, previous work~\cite{weston2019automated, wang2017two} focused on body composition analysis, segmenting the image into different body structures, including subcutaneous adipose tissue and the external skin edge. In~\cite{ognard2019edge}, a combination of the Canny filter, the selection of boundaries and a local regression has been applied to delimitate the different skin layers in 3T MRI with T2-weighted sequence. All these works have developed skin segmentation as part of their work pipelines, thus focusing on one imaging modality and leveraging the properties of that specific image.
	
\paragraph*{3D Rigid registration}
3D rigid registration refers to aligning two 3D surfaces or point clouds. The \emph{Iterative Closest Point} (ICP) algorithm~\cite{wang2017survey} iteratively searches the closest points between two point clouds and computes a rigid transformation to align them. The \emph{Robust Point Matching} (RPM)~\cite{rangarajan1997robust} applies a probabilistic approach to estimate the correspondences between points in two 3D point clouds. It is less sensitive to noise and outliers than ICP. The \emph{Coherent Point Drift} (CPD)~\cite{myronenko2010point} applies a Gaussian mixture model to model the probability distribution of the point clouds. It supports rigid and non-rigid deformations and is more versatile than ICP and RPM. Deep learning methods, such as PointNet~\cite{aoki2019pointnetlk, qi2017pointnet} and PointNet+~\cite{qi2017pointnet}, apply neural networks to learn features from 3D point clouds and perform registration. Deep learning methods are highly efficient regarding the time required for registration after the training, thus valuable for real-time applications. However, learning methods must be trained on large data to avoid biases, and having large and various data sets in medical applications is still a challenge.

\paragraph*{3D Sensors for medical applications}
The value of 3D camera trackers, depth sensors, or LiDAR scanners has been firmly established across various fields. Medical imaging, radiology, and surgery are some domains that can benefit significantly from their integration. In recent years, extensive research efforts have been directed towards enhancing the resolution, accuracy, and speed of these sensor technologies~\cite{sakas2002trends}. Leveraging the data captured by 3D sensors holds the potential for substantial improvements in several medical applications. These advancements span diverse areas, ranging from surgical navigation and robot-assisted surgery, where 3D sensors facilitate image-guided procedures and enable robots to assist surgeons, to the domain of rehabilitation and physical therapy, where these sensors are already being employed to monitor patient movements meticulously and offer valuable feedback to both patients and therapists. Radiology and Imaging, in particular, emerge as a pivotal field poised to leverage the capabilities of 3D sensors ~\cite{von2021medical}. Their integration enhances the quality of medical images, including those from CT and MRI scans, by providing real-time feedback on patient positioning and motion throughout the imaging process. Within this area, our research is centred on the challenge of leveraging a 3D camera tracker in radiology, which is necessary for harmonising heterogeneous data sources, notably the volumetric pre-operative image and the 3D surface data derived from the camera.
	%
	%\begin{figure}[t]
	%\centering
	%\includegraphics[width=0.8 \linewidth] { /systemPic}
	%\caption{\label{fig:completeSystem}Image fusion system setup comprehensive of the tracking system and the software registration input rendering (i.e. the segmented surface and the mesh acquired by the 3D camera).} 
	%\end{figure}
	%
	%\subsection{MR~$\&$ US fusion: overview and main features\label{sec:overview_method}}
	%
	\begin{figure}[t]
		\centering
		\centering
		\begin{tabular}{ccc}
			(a)\includegraphics[width=0.4 \linewidth]{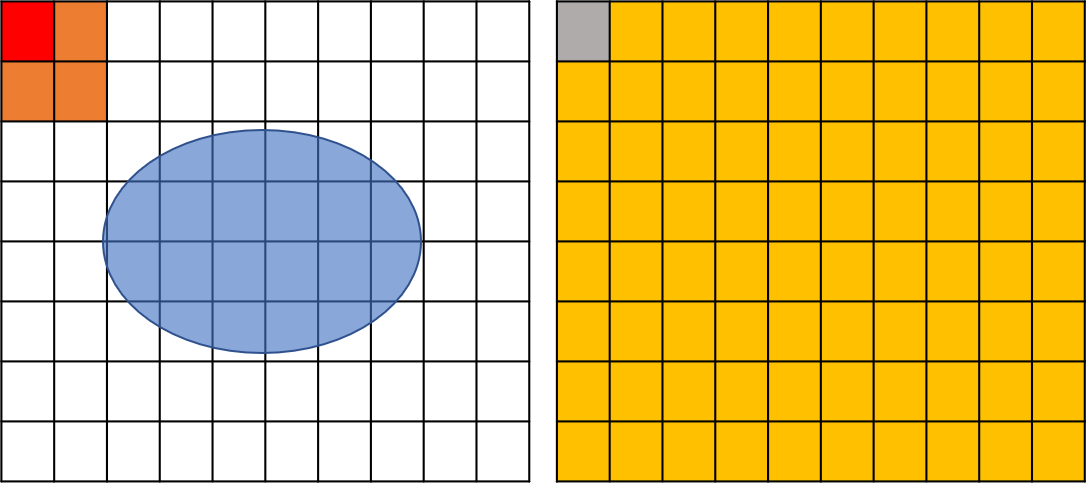}&
			(b)\includegraphics[width=0.4 \linewidth]{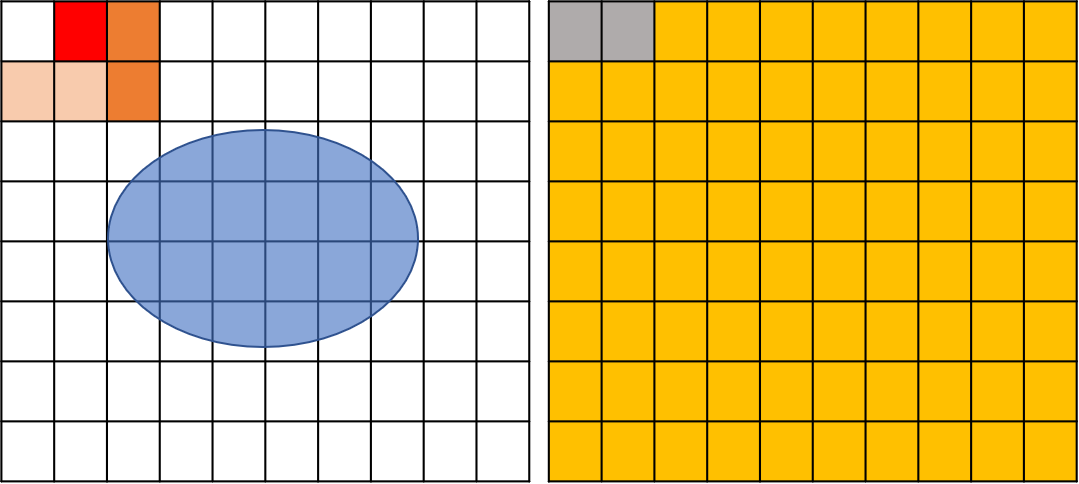}\\
			(c)\includegraphics[width=0.4 \linewidth]{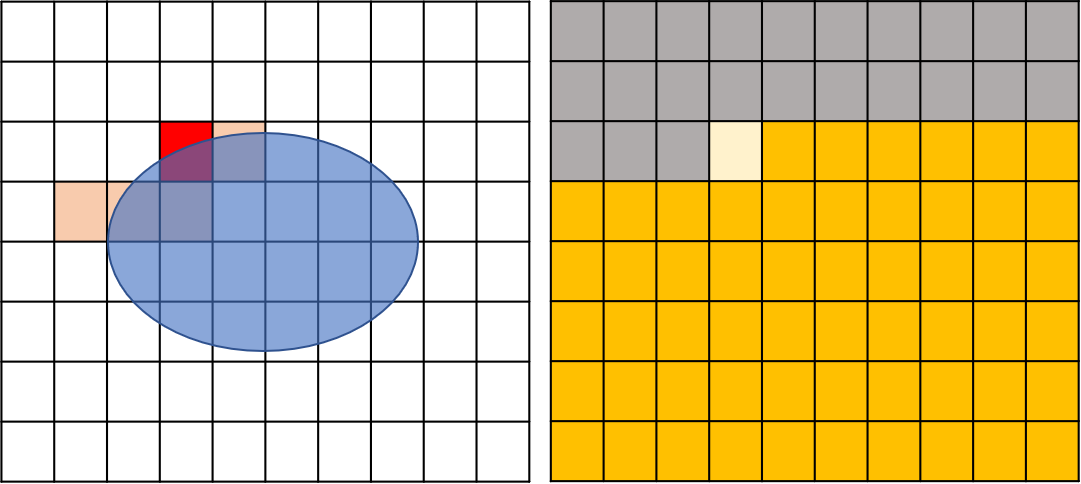}&
			(d)\includegraphics[width=0.4 \linewidth]{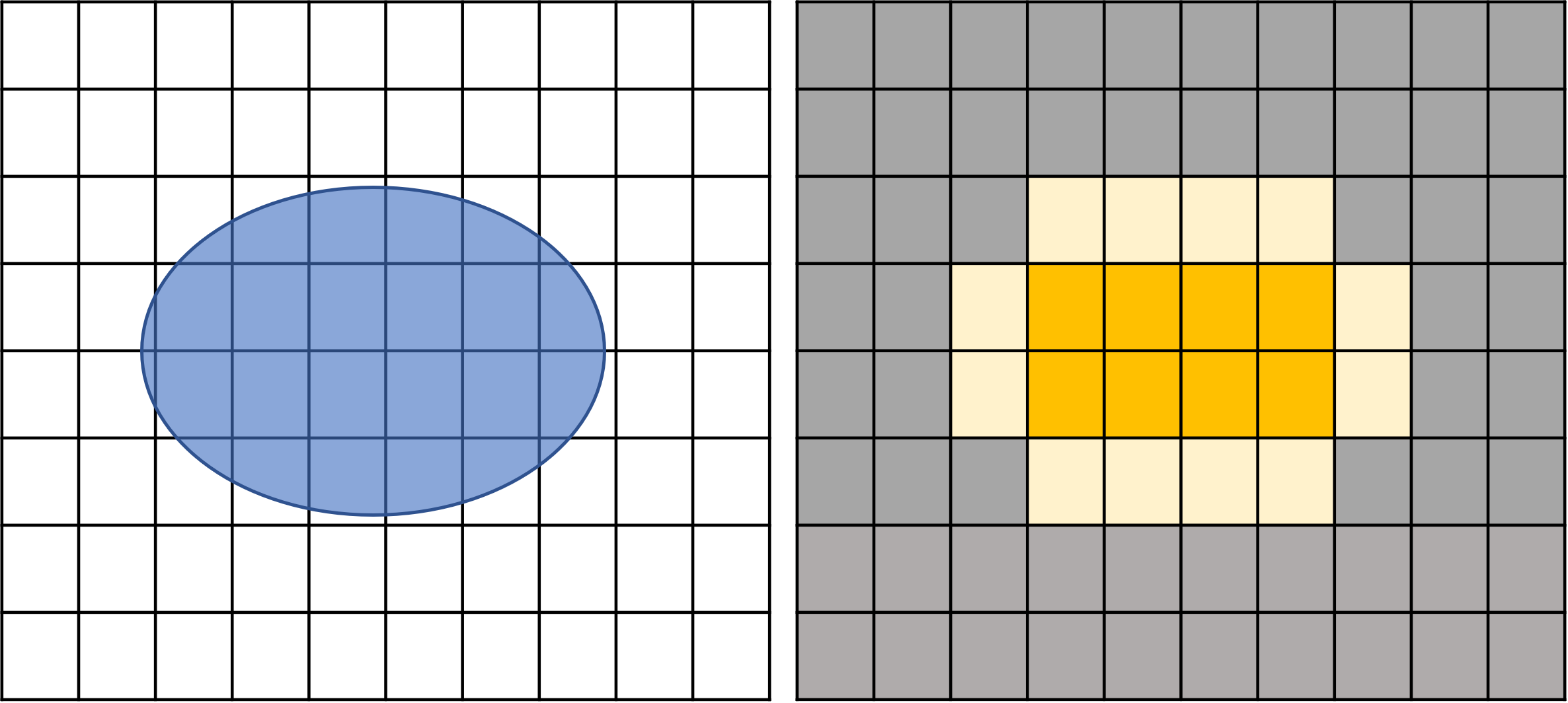}
		\end{tabular}
		\caption{\label{fig:mockup}Description of the segmentation method on a slice. The toy image is on the left of each step, and the mock-up grid is on the right. The red pixels are under evaluation, and the orange pixels are inserted in the "visited pixel" list by the current evaluation step. The lighter orange pixels are inserted in the list by previous steps. (a) The algorithm's initialisation (every pixel in the mock-up gridhas a value of 2), evaluation of the first pixel, identification of the neighbourhood and consequent assignment of background value on the mock-up grid. (b) The second step of the algorithm has the same consideration as the previous one. (c) Identification of a pixel above the threshold: the value of the mock-up pixel changes to 1, and the neighbourhood of the pixel is not inserted in the list. (d) Final segmentation.}
	\end{figure}
\section{MR~$\&$ US fusion system\label{sec:PROPOSED-APPROACH}}
	\begin{figure}[t]
		\centering
		\centering
		\begin{tabular}{ccccccc}
			(a)\includegraphics[width=0.3 \linewidth]{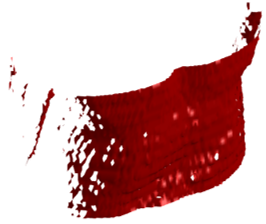}&
			(b)\includegraphics[width=0.3 \linewidth]{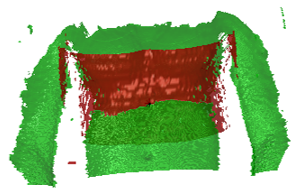}\\
			(c)\includegraphics[width=0.3 \linewidth]{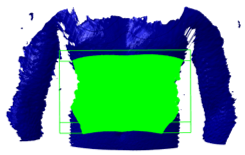}&
			\includegraphics[width=0.3 \linewidth]{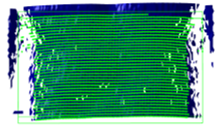}\\
			(d)\includegraphics[width=0.3 \linewidth]{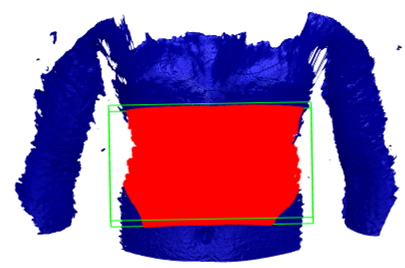}&
			\includegraphics[width=0.3 \linewidth]{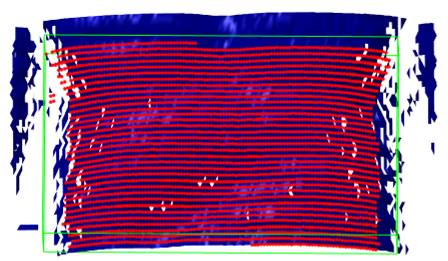}&
		\end{tabular}
		\caption{\label{fig:registrationPipeline} Co-registration steps. (a) The selected anterior portion of the segmented surface. (b) PCA alignment and translation on the reference points. (c) Surface sub-regions for the first ICP run. (d) Surface sub-regions tuning for the second ICP step (only $80\%$ of the surface considered in the first ICP run is considered in the second refinement run).}
	\end{figure}
	\begin{figure}[t]
		\centering
		\begin{tabular}{cccc}
			(a)\includegraphics[width=0.25 \linewidth]{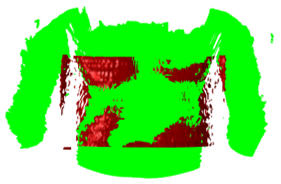}&
			(b)\includegraphics[width=0.25 \linewidth]{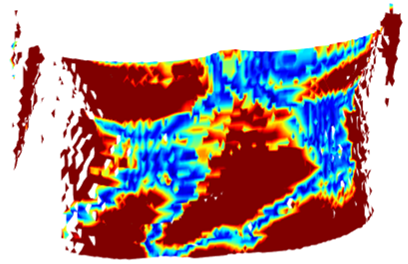}&
			(c)\includegraphics[width=0.25 \linewidth]{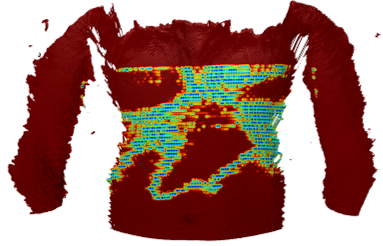}&
			\includegraphics[height=0.15 \linewidth]{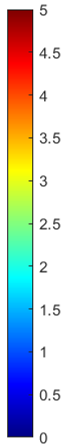}
		\end{tabular}
		\caption{\label{fig:registrationResult} (a) Co-registration pipeline result. Error distribution (b) on the segmented surface and (c) on the camera mesh. The unit measure of the colourmap is mm.}
	\end{figure}
	
The novel co-registration is divided into two main software components integrated into the image fusion system. The first aims to segment the external skin surface of patients acquired by MR/CT, while the second seeks to co-register the segmented skin surface with the 3D surface obtained by the camera.
	
\paragraph*{Skin segmentation\label{sec:SKIN-SEG-OVERALL}}
The segmentation of the 3D surface representing the patient's skin is used to bridge the seamless integration of the heterogeneous data sources involved in the system. Indeed, extracting the body surface from volumetric imaging facilitates subsequent analyses and enables the processing of lighter data. The segmentation of the external body surface is computed according to the Hounsfield value (CT) or intensity level (MR), set as a default parameter, and is represented as triangle mesh. Given a CT/MR image paired with the skin iso-value, the proposed segmentation identifies the subject's skin surface, which is used as input for the co-registration. The general idea is to leverage the differences in intensities between the air and the body surface. The segmentation proceeds one slice at a time, starting from a background pixel. Then, the growth of the background region proceeds iteratively based on the pixels’ adjacency and stops when it encounters a pixel whose grey value is higher or equal to the iso-value of the skin. Through this region-growing algorithm, the evaluation expands only where the air is present; the body will be segmented as a whole object by exclusion.

	Fig.~\ref{fig:mockup} describes the segmentation method developed. We create a mock-up grid of the same dimensions of each slice, and each element of the grid is related to the pixel in the original slice that presents the same location (Fig.~\ref{fig:mockup}(a)). Initially, all the grid elements of the mock-up are associated with the same \textit{initial value} equal to 2. On the image slice, the starting pixel must belong to the background, e.g., the pixels in the corners, since the subject’s body is typically located in the centre of the image. Once the starting pixel has been selected, we check if it belongs to the background through its intensity level in the original slice. A pixel is considered background if its intensity remains under the skin isovalue in the image. In contrast, if its intensity is above the isovalue, it indicates that we encountered the body edge.
	
	If the pixel belongs to the background, the corresponding element in the mock-up grid is associated with 0, and the pixel is marked as visited. Then, we select its neighbouring pixels, which initialise the list of pixels to be subsequentially visited (Fig.~\ref{fig:mockup}(b)). Iteratively, we check if the first pixel of the list belongs to the background. If it belongs in the background, we proceed as before and update the list by removing the just visited pixel, marking it as visited and adding its neighbourhood. Pixels that have already been visited before must not be inserted in the list again (Fig.~\ref{fig:mockup}(c)). If the pixel in the original image is above the skin isovalue, then we assign the value of 1 to the corresponding element in the mock-up grid.
	Moreover, we do not insert the pixel's neighbourhood in the list (Fig.~\ref{fig:mockup}(d)). At the end of the process, the mock-up grid presents a value of 0 in the background, a value of 2 inside the subject's body (i.e. the initial value of all the elements in the mock-up grid) and 1 in correspondence with the skin. The same procedure is applied to all the slices in the volume.
	
	Then, the segmented volume undergoes the marching cube algorithm~\cite{lorensen1987marching} to extract a 3D surface mesh of the segmented skin, representing the input for the co-registration phase to match the MRI and the 3D surface acquired by the camera. To improve the segmentation, we add padding around each slice coloured with the minimum value appearing in the image to guarantee that it will be considered background. In this way, the algorithm proceeds through the padding pixels toward the slice's end. The padding is helpful in case the MR/CT bed has been acquired with the patient. To reduce the overall computational time for skin segmentation, we can sub-sample each slice and each set in the case of high-resolution MR/CT images. The intensity value for the skin, i.e. the iso-value needed as an input parameter to the segmentation algorithm, is easily retrievable according to the specification of the MR/CT acquisition machine. Indeed, usually, each manufacturer has standard values for each imaging modality.
	\begin{figure}[t]
		\centering
		\centering
		\begin{tabular}{ccc}
			(a)\includegraphics[width=0.25\linewidth]{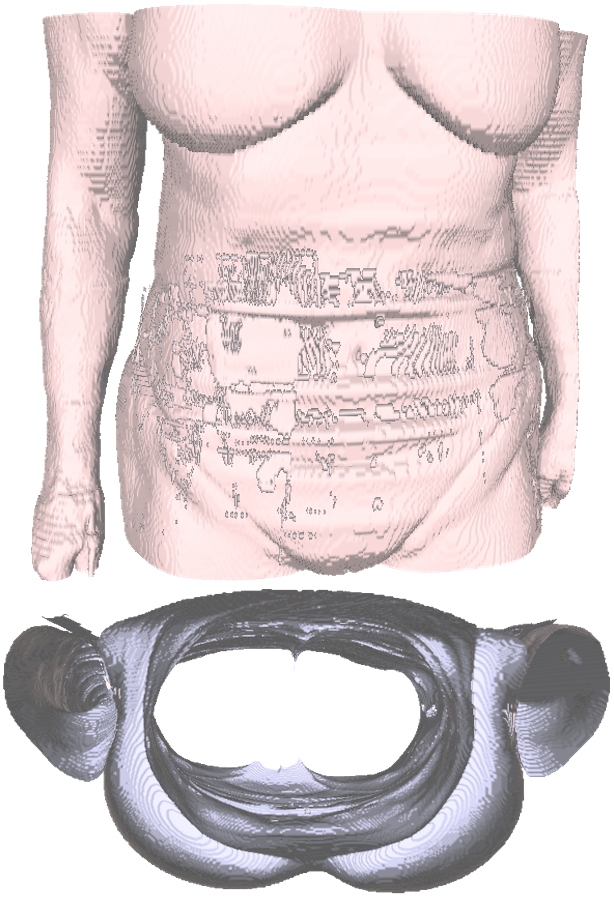}&
			(b)\includegraphics[width=0.25 \linewidth]{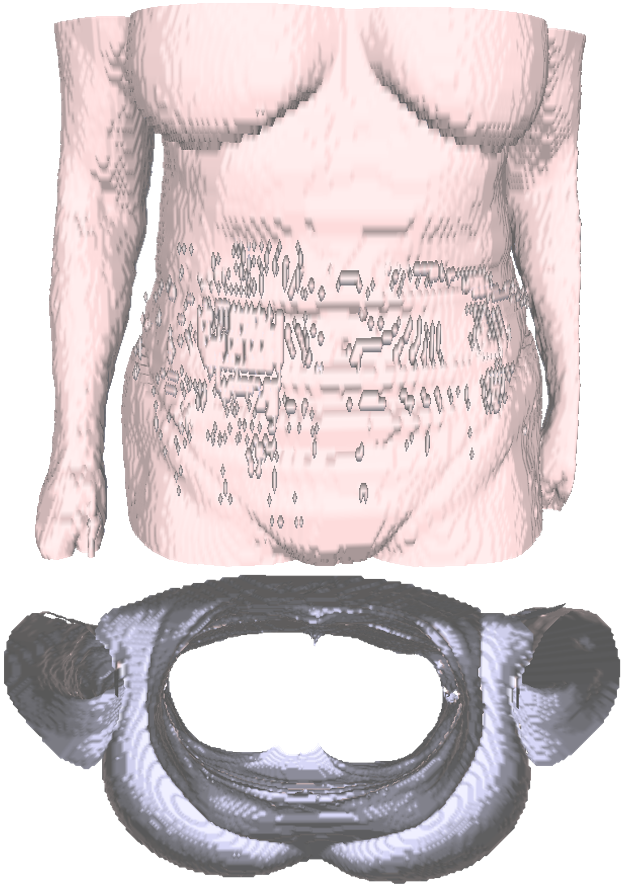}&
			(c)\includegraphics[width=0.35 \linewidth]{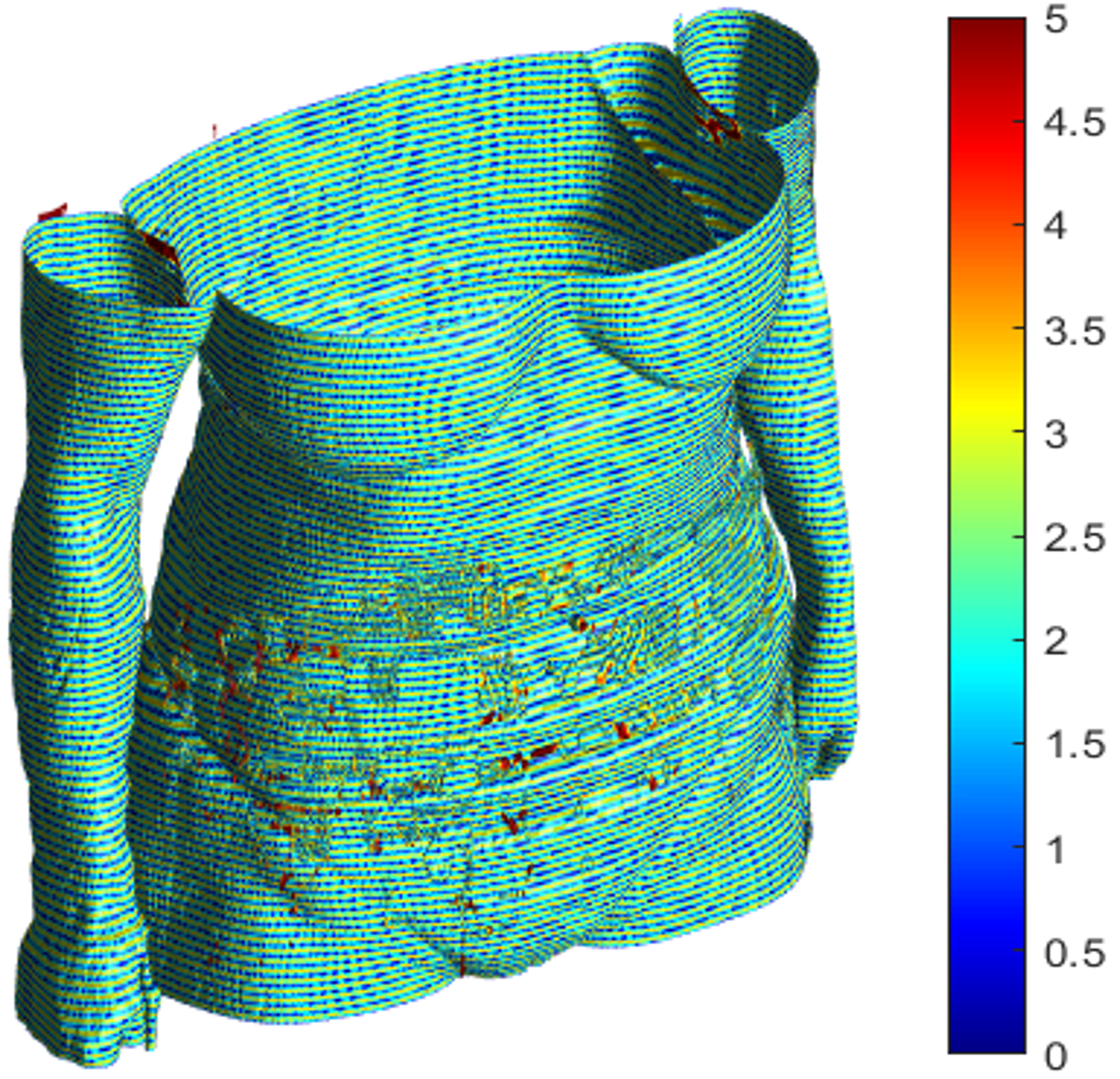}
		\end{tabular}
		\caption{\label{fig:subsample}Robustness to image subsampling of the skin segmentation method. Skin surface extracted from the segmentation of the input image at (a) high and (b) low resolution. (c) Distance distribution between the two surfaces; the colourmap scale goes from~$0$ mm (blue) to~$5$ mm (red). }
	\end{figure}
	\begin{figure}[t]
		\centering
		\begin{tabular}{cccc}
			%(a)\includegraphics[height=0.25 \linewidth]{faccia.png}&
			(a)\includegraphics[height=0.25 \linewidth]{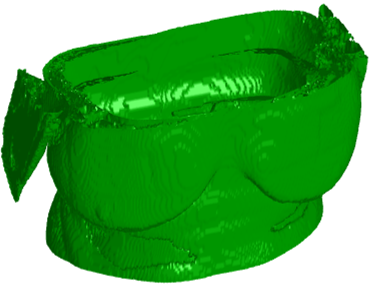}&
			(b)\includegraphics[height=0.25 \linewidth]{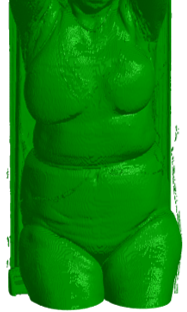}
		\end{tabular}
		\caption{\label{fig:segmDistrict}Skin segmentation of (a) a head MRI pipeline, (b) an abdomen MRI, (c) a whole-body CT.} 
	\end{figure}

	%
	%\begin{figure}
	%\centering
	%\includegraphics[width=0.5\textwidth]{pipeline.png}
	%\caption{\label{fig:pipeline}Pipeline of the segmentation algorithm.}
	%\end{figure}
	%
	%\begin{figure}[t]
	%\centering
	%\centering
	%\begin{tabular}{ccc}
	%(a)\includegraphics[width=0.4 \linewidth]{mockupA.png}&
	%(b)\includegraphics[width=0.4 \linewidth]{mockupB.png}\\
	%(c)\includegraphics[width=0.4 \linewidth]{mockupC.png}&
	% (d)\includegraphics[width=0.4 \linewidth]{mockupD.png}
	%\end{tabular}
	%\caption{\label{fig:mockup} Description of the algorithm on a slice. The toy image is on the left of each step, and the mock-up grid is on the right. The red pixels are under evaluation, and the orange pixels are inserted in the "visited pixel" list by the current evaluation step. The lighter orange pixels are inserted in the list by previous steps. (a) The algorithm's initialisation, evaluation of the first pixel, identification of the neighbourhood and consequent assignment of background value on the mock-up grid. (b) The second step of the algorithm has the same consideration as the previous one. (c) Identification of a pixel above the threshold: the value of the mock-up pixel does not change, and the neighbourhood of the pixel is not inserted in the list. (d) Final segmentation.}
	%\end{figure}
	%
	%\subsubsection{3D skin segmentation\label{sec:SKIN-SEG}}
	%
	\begin{figure}[t]
		\centering
		\begin{tabular}{cccccccc}
			(a)\includegraphics[width=0.2\linewidth]{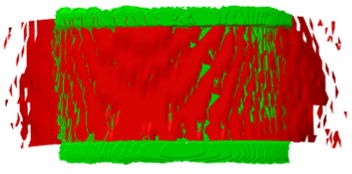}&
			(b)\includegraphics[width=0.2\linewidth]{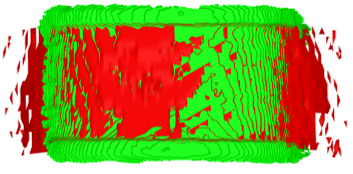}&
			(c)\includegraphics[width=0.2\linewidth]{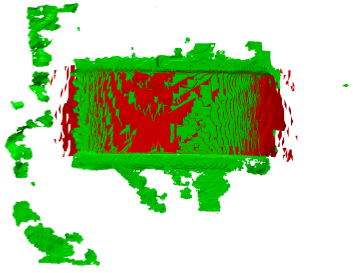}&
			(d)\includegraphics[width=0.2\linewidth]{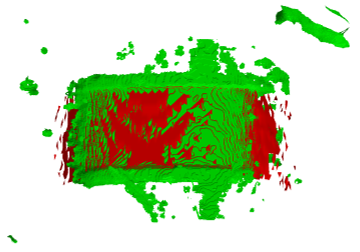}\\
			(e)\includegraphics[width=0.2\linewidth]{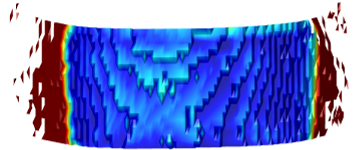}&
			(f)\includegraphics[width=0.2\linewidth]{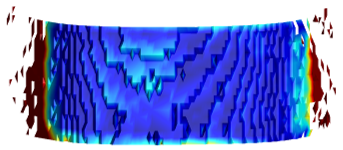}&
			(g)\includegraphics[width=0.2\linewidth]{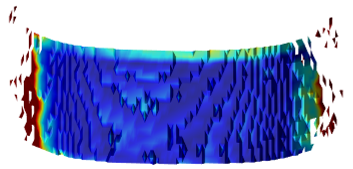}&
			(h)\includegraphics[width=0.2\linewidth]{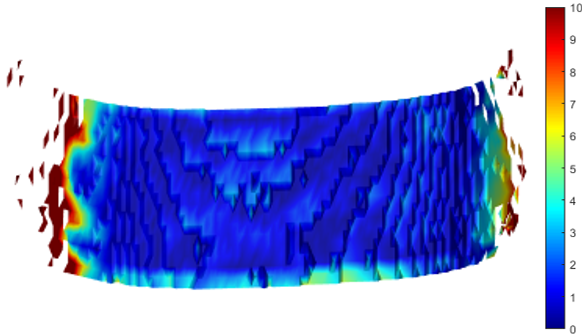}&
		\end{tabular}
		\caption{\label{fig:distRobust} Co-registration and error distribution on a phantom acquired by a 3D camera at different distances: (a,e)~$20$ cm, (b,f)~$25$ cm, (c,g)~$30$ cm, (d,h)~$35$ cm. The error visualisation scale in the colour map goes from~$0$ mm (blue) to~$10$ mm (red).} 
	\end{figure}
\paragraph*{Skin co-registration}\label{sec:REGISTRATION}
To align the MR/CT image with the US probe, the 3D surface acquired by the camera, which is in the same reference system as the US probe and of the magnetic tracking, is rigidly co-registered with the patient skin segmented from the MR/CT images. The output of the co-registration is a translation vector and a rotation matrix, which co-register the segmented surface to the 3D surface acquired by the camera (i.e., the Intel RealSense in the experimental setup), minimising the corresponding misalignment. The co-registration takes the segmented surface extracted from the anatomical images (MRI/CT), the 3D surface acquired by the camera and a reference virtual landmark as inputs. The 3D surface must be acquired by the 3D camera with a frontal view, following the guidance provided by the camera to minimise acquisition errors. The segmented surface is oriented consistently (i.e., head-feet, right-left) to avoid errors related to body symmetries. The operator manually selects one corresponding landmark point on each input surface to align the two surfaces through an intuitive interface. Thus, the landmark is exclusively virtual and does not require any external physical placement. A pipeline composed by orientation adjustment through \emph{Principal Component Analysis} (PCA), surface sub-region selection and tuning (region of interest), and various coregistration refinements leveraging the \emph{Iterative Closest Point algorithm} (ICP) allows for accurate alignment of the segmented surface with the surface acquired by the camera (Fig.~\ref{fig:registrationPipeline}). Then, the computed roto-translation is applied to the volumetric data to put the MR/CT images in the same reference system of the US probe, thus enabling the fusion of the MR/CT image with the US image since the tracking system tracks both the 3D camera and the US probe.
	
	This result will allow the radiologist/surgeon to navigate the MR and US images simultaneously during the US examination or preoperatively. To optimise the time to rigidly register the segmented surface with the 3D surface acquired by the camera, the segmented surface is cut to get only the front part of the body. This way, the relevant part of the segmented surface undergoes the registration. To cut the surface, we consider the angle between the normal at each surface vertex and the sagittal axis: the vertices associated with an angle smaller than 90\textdegree are selected as part of the front surface.
	\begin{figure}[t]
		\centering
		\begin{tabular}{cccccccc}
			\includegraphics[width=0.27\linewidth]{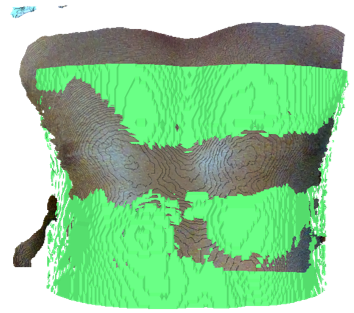}&
			\includegraphics[width=0.3\linewidth]{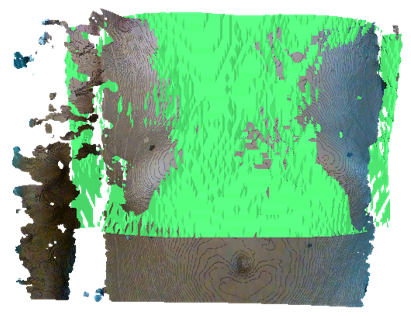}&
			\includegraphics[width=0.3\linewidth]{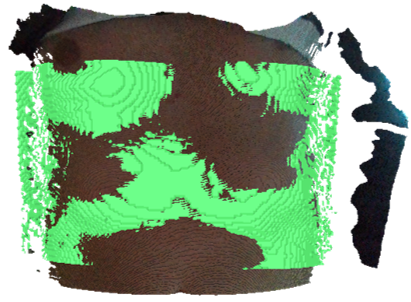}&
		\end{tabular}
		\caption{\label{fig:resultwithtexture} Co-registration between the (textured) surface acquired by the camera and the segmented surface (green) on different subjects.}
	\end{figure}

Visualising the registration error between the segmented and acquired skin (Fig.~\ref{fig:registrationResult}) gives valuable insight to assess whether acquiring a more accurate surface from the camera is necessary to improve the co-registration or if the (co-registration) results are accurate enough. The co-registration error between the segmented skin and the skin acquired by the camera is computed as the Hausdorff distance between the co-registered surfaces. Calling the segmented surface~$\mathbf{X}_{1}$ and the 3D surface acquired by the camera ~$\mathbf{X}_{2}$ we identify co-registration error by computing their Hausdorff distance~$d(\mathbf{X}_{1},\mathbf{X}_{2}):=\max\{d_{\mathbf{X}_{1}}(X_{2}),d_{\mathbf{X}_{2}}(X_{1})\}$. Where~$ d_{\mathbf{X}_{1}}\left( \mathbf{X}_{2}\right) :=\max _{\mathbf{x}\in \mathbf{X}_{1}}\left\{ \min _{\mathbf{y}\in \mathbf{X}_{2}}\left\{ \left\|\mathbf{x}-\mathbf{y}\right\| _{2}\right\} \right\}~$. The minimum distance is calculated using a Kd-tree structure. Higher co-registration errors present a higher Hausdorff distance. The distance distribution is mapped to RGB colours, and each vertex is assigned the corresponding colour according to its distance from the other surface. To better analyse the distance distribution in the relevant portion of the surface, vertices that present a distance equal or higher than 5mm are coloured as red. Vertices that have a null distance are shown in blue. The other distances are mapped to the shades between red and blue. If the error is located in areas relevant to the structure under analysis, it may prompt reconsidering the data acquisition process. Conversely, if errors are primarily present in regions not critical to the examination, the medical professional can confidently proceed with analysing the fused MR/CT and US images.
	\begin{figure}[t]
		\centering
		\begin{tabular}{cc}
			\includegraphics[width=0.46\linewidth]{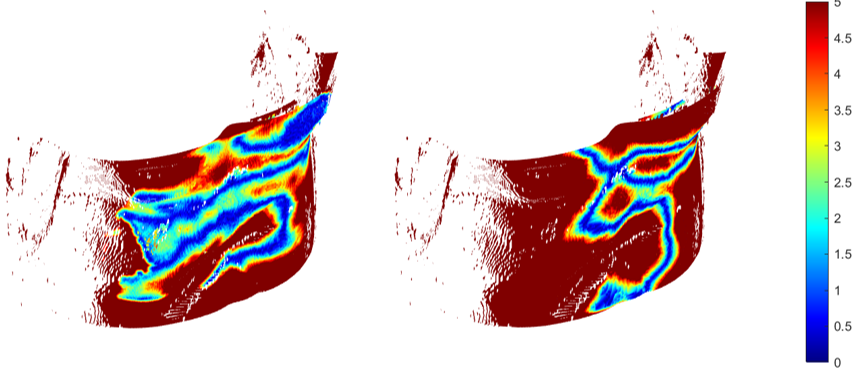}
			&\includegraphics[width=0.46\linewidth]{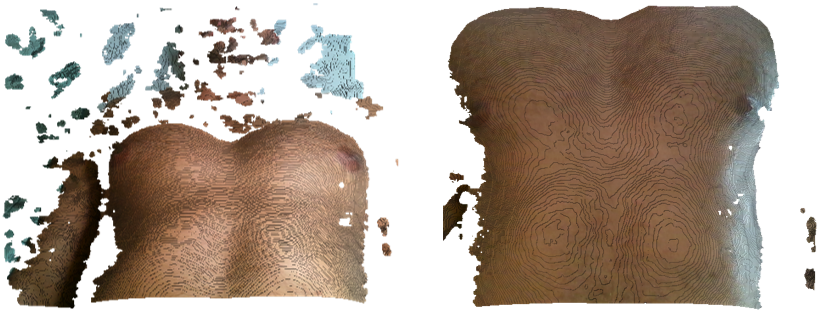}
		\end{tabular}
		\caption{\label{fig:resulinclination} Co-registration error on the same subject with different camera view angles: left 45 degrees, right 90 degrees. The colourmap scale goes from~$0$ mm (blue) to~$5$ mm (red). The acquisition quality becomes less accurate when the camera is kept at 90 degrees with respect to the subject (right). In comparison, it becomes more accurate if the camera is inclined 45 degrees as specified in the datasheet (left).} 
	\end{figure}
	\section{Experimental results and validation}\label{sec:RESULTS}
	We discuss the results on skin segmentation, the robustness of the skin co-registration to noise and selected parameters (e.g., HU value, virtual landmark), and the accuracy of the image fusion. 
	\begin{figure}[t]
		\centering
		\begin{tabular}{ccc}
			\includegraphics[width=0.30\linewidth]{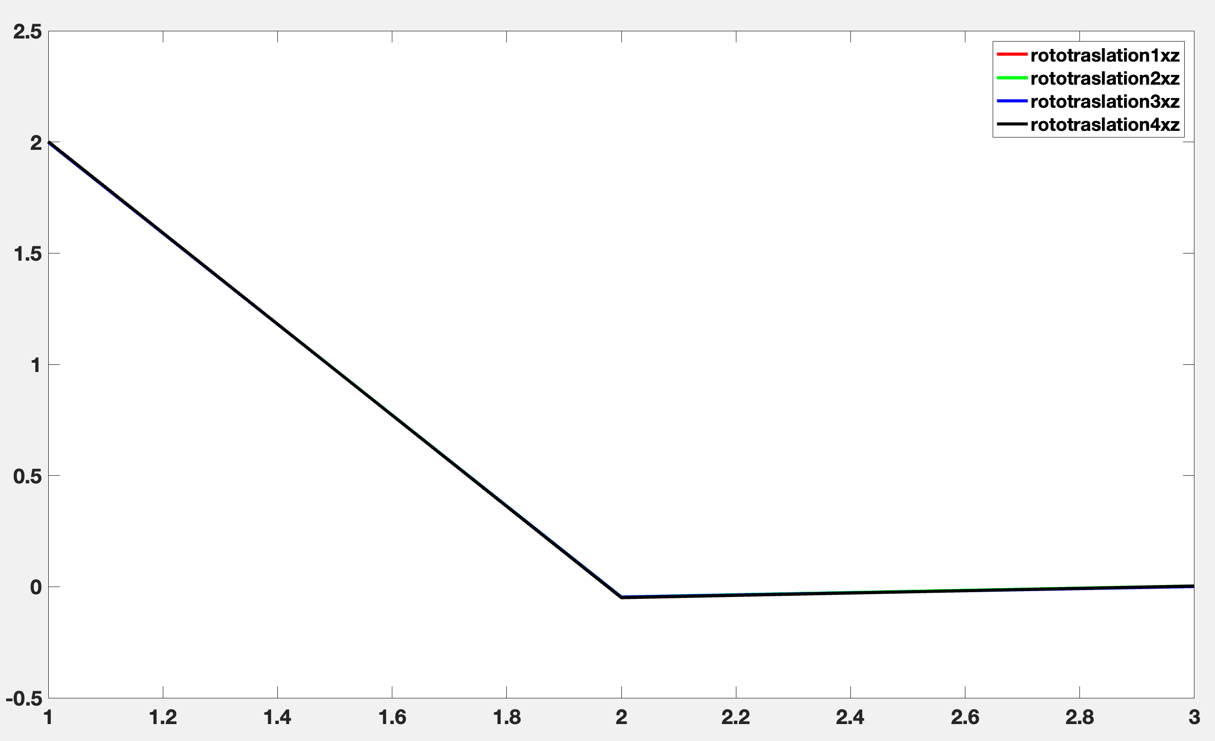}&
			\includegraphics[width=0.30\linewidth]{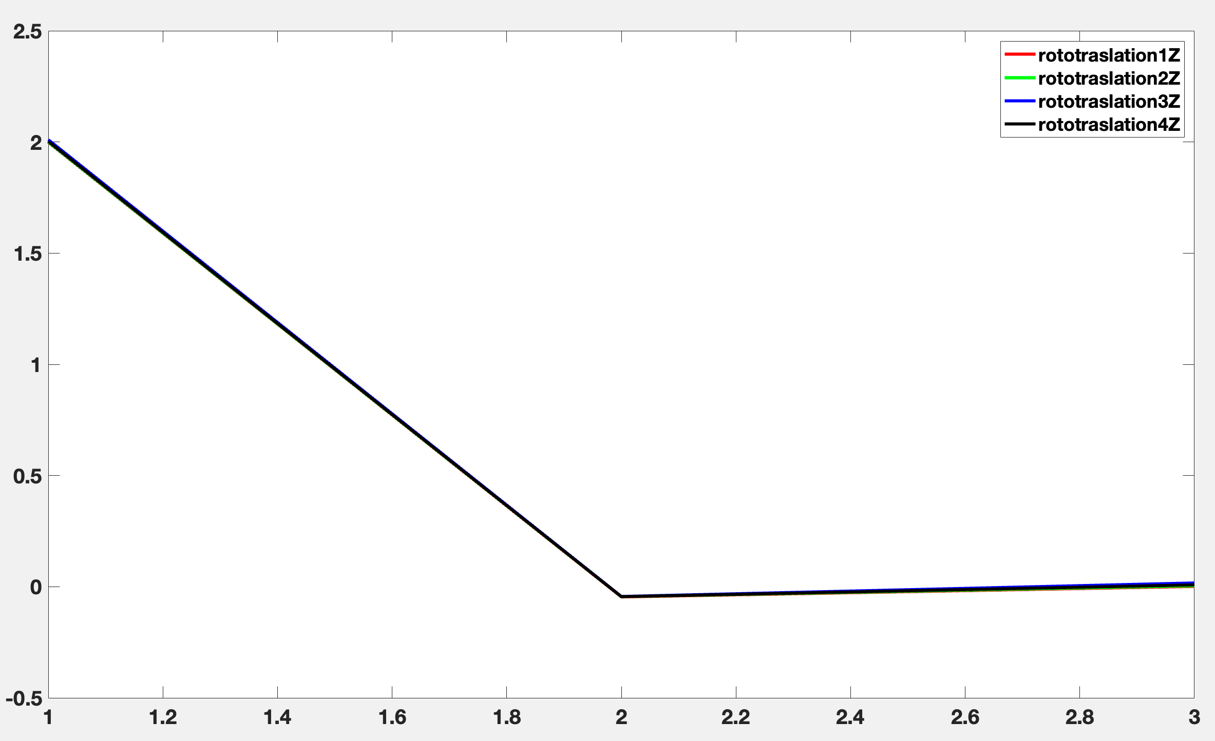}&
			\includegraphics[width=0.30\linewidth]{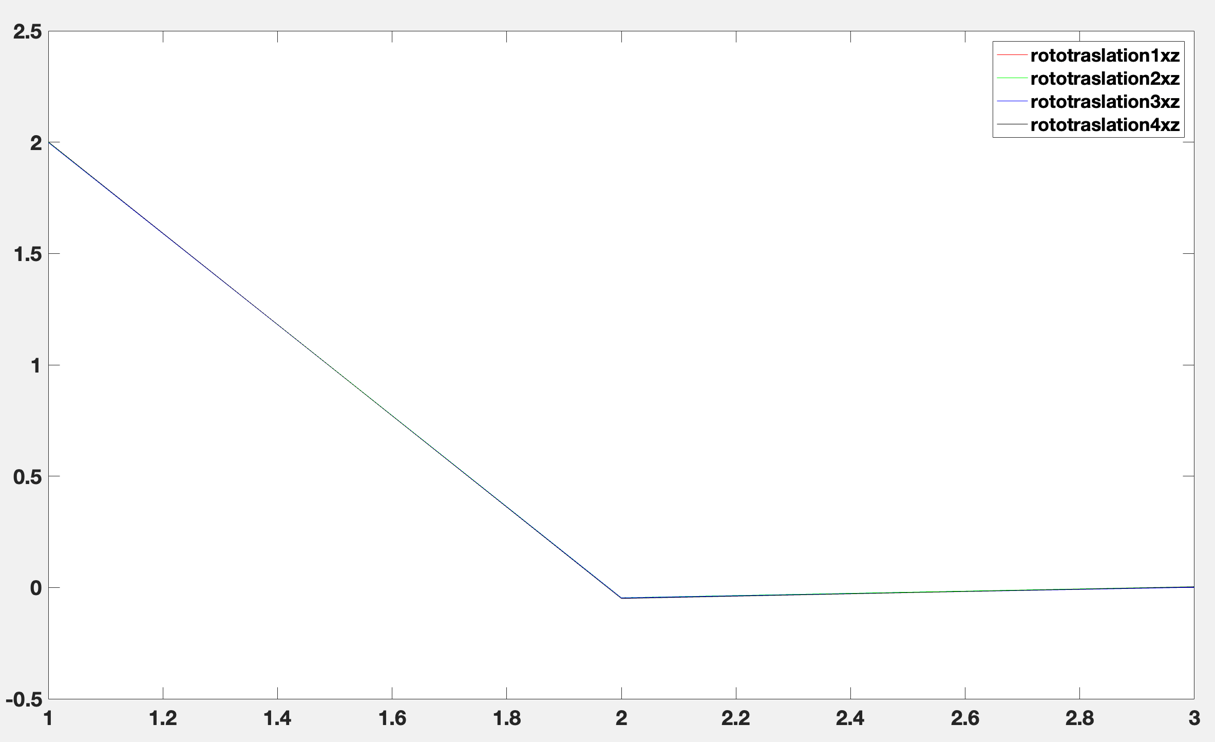}\\
			(a) &(b) &(c)
		\end{tabular}
		\caption{\label{fig:resultMarker} Rotation angles concerning the~$X$,~$Y$, and~$Z$ axes when a misalignment on the marker selection is present. (a) the misalignment is in the~$X$ direction, (b) in the~$Z$ direction and (c) in a diagonal direction ($X$ and~$Y$). The rotation angles remain the same. Thus, the algorithm is robust to errors in the virtual landmark selection.} 
	\end{figure}
	%
	%\subsection{Experimental results\label{sec:EXP-RESULTS}}
%
\paragraph*{Skin segmentation}\label{sec:SKIN-RESULTS}
The segmentation (i.e., the voxel labelling) and mesh extraction have a computational cost linear to the number of voxels composing the volumetric image. Table~\ref{tab:timeTable} reports the timing of each algorithm step on an 11th generation Intel(R) Core(TM) i7-11700k 8 core. To better integrate the approach with existing clinical workflow, we tested its robustness to subsampling. Indeed, given the computational cost property of the method, even a light subsampling could drastically improve the performance in terms of time required for the segmentation to complete. Fig.~\ref{fig:subsample} shows how skin segmentation remains clean and accurate given an image and its subsampled version. We subsampled a volume image by a factor of two in each direction. The only difference is in the resolution of the surface, which is a direct consequence of the lower resolution of the subsampled original image. To confirm the maintained accuracy of the segmentation at different image resolutions, we computed the distance distribution between the surfaces extracted from a volume image and its subsampled version. In this case, the higher distances correspond to those slices and pixels missing in the subsampled version of the image, and the value of the distance is coherent with the changed dimension of the voxels. 
The 3D skin segmentation, contrary to AI methods, does not require any training and consequently does not require a large data set or various acquisition of diverse imaging modalities, contributing to the method's generality. 
The skin segmentation has been designed to be as general as possible regarding the anatomical area scanned (e.g., head, breast, total body and abdomen) and the acquisition modality (e.g., MR, CT, PET). We tested the quality of the segmentation on different anatomical volume images, such as CTs and MRIs (Fig.~\ref{fig:segmDistrict}), obtaining satisfactory results.
	\begin{table}[t]
		\centering
		\caption{Computing time of 3D skin segmentation algorithm on various anatomical districts and imaging modalities.}
		\begin{tabular}{|l l l l l l|}
			\hline
			\textbf{Imaging} & \textbf{Volume} & \textbf{District} & \textbf{Volume} & \textbf{Segmentation} & \textbf{Skin extraction}\\
			\textbf{Modality} & \textbf{size} &  & \textbf{reading} &  & \textbf{extraction}\\
			\hline
			MRI T1 &~$260\times 52\times 72$ & Abdomen &~$3$s &~$28$s &~$4$s\\ 
			\hline
			MRI T2 &~$184\times 256\times 30$ & Abdomen &~$1$s &~$5$s &~$0.2$s\\
			\hline
			MRI &~$384\times 384\times 50$ & Breast &~$3$s &~$45$s &~$3$s\\
			\hline 
			MRI T2 &~$384\times 384\times 46$ & Breast &~$2$s &~$42$s &~$2$s \\
			\hline
			CT &~$512\times 512\times 247$ & Whole body &~$2$s &~$304$s &~$16$s \\ 
			\hline
			CT & ~$256\times 256\times 160$ & Head &~$1$s &~$46$s &~$1$s \\ 
			\hline
		\end{tabular}
		\label{tab:timeTable}
	\end{table}
\paragraph*{Co-registration\label{sec:COREG-RESULTS}}
The co-registration experimental tests were performed on both the phantom of the abdomen and real subjects. The error on a phantom is related mainly to the camera position, the acquired area's limited dimension, and its symmetric shape. The accuracy inside the volume with the phantom tests is satisfactory since the error remains within an acceptable range (i.e., under 1 cm), and the physician can quickly correct the smaller error through a fast tuning performed manually to produce a more accurate result (if needed). To verify the robustness of the co-registration, the skin surface was captured by placing the 3D camera at different distances from the skin (Fig.~\ref{fig:distRobust}). The co-registration remains stable against the acquisition distance. At very high distances, the noise acquired by the camera increases, confirming the algorithm's robustness to the noise and symmetries, which are typical of the phantom but unlikely in real subjects. Fig.~\ref{fig:resultwithtexture} shows the co-registration on CT/MRI images of real subjects. We tested how much tilting the 3D camera at various angles during the acquisition would affect the co-registration. From the camera's specifications, the ideal acquisition angle is 45 degrees to avoid distortions in the acquired 3D surface. Fig.~\ref{fig:resulinclination} shows the co-registration when the camera is tilted by 45 degrees, which has the best co-registration due to the higher quality of the acquired surface, and when the camera view is perpendicular to the patient's surface. The additional noise with a view angle of 45 degrees does not interfere with the co-registration, which includes the automatic identification of regions of interest both on the segmented surface and the 3D surface acquired by the camera. This region of interest identifies correspondent regions on the two surfaces and notably reduces the inclusion of noise from the 3D camera. To verify the influence of the selection of corresponding virtual landmarks on the co-registration, we select slightly displaced markers along the~$X$, the~$Z$ axes and a diagonal ($X$ and~$Y$ displacement) directions and measure the changes of the angles between the corresponding co-registration matrices. Considering increasing displacements on the landmark selection, significant differences in the rotation angles were not found (Fig.~\ref{fig:resultMarker}). The robustness of the co-registration to a misplacement of the reference virtual landmarks confirms that the algorithm will not be user-dependent, i.e., the users can apply different approaches to select the landmark.
	\begin{figure}[t]
		\centering
		\begin{tabular}{cccc}
			(a)\includegraphics[width=0.8\linewidth]{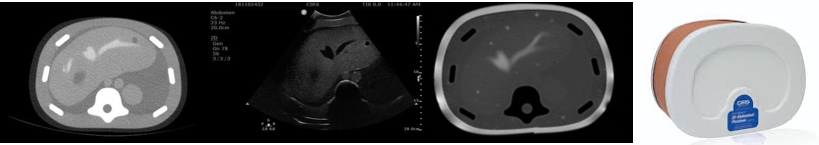}\\
			(b)\includegraphics[width=0.8\linewidth]{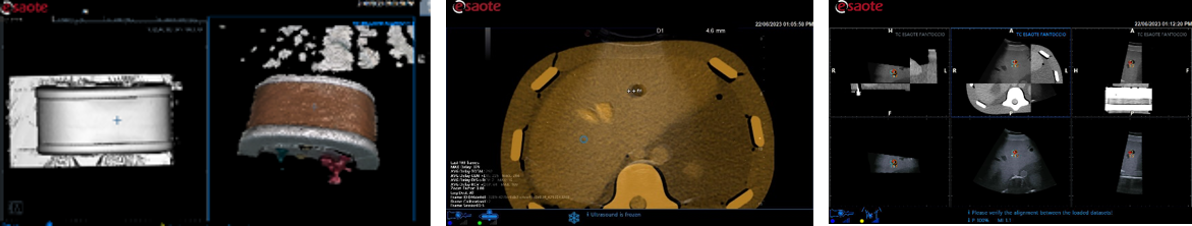}\\
		\end{tabular}
		\caption{\label{fig:result0}(a) MRI image of a phantom of the abdominal district.(b) From left to right: Phantom surface acquired at 0 degrees, image fusion results with CT, and accuracy error 4.3 mm.}
	\end{figure}
	%
 %(Fig.~\ref{fig:phantomTest}) 
	%
	\begin{figure}[t]
		\centering
		\includegraphics[width=0.5\linewidth]{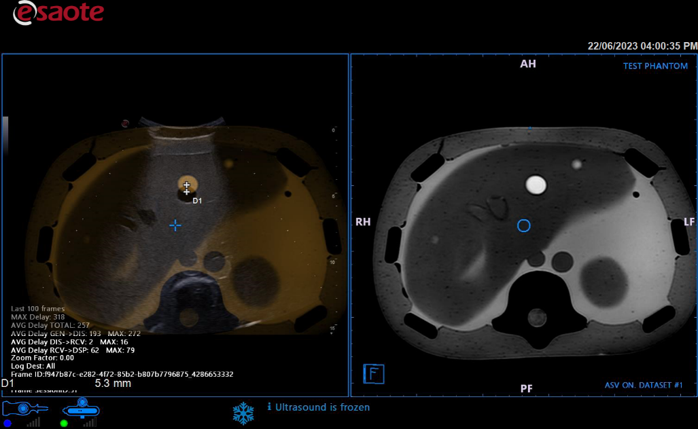}
		\caption{\label{fig:result0MRI}US co-registration with an MR: accuracy error 5.3 mm. The skin surface has been acquired at 0 degrees.} 
	\end{figure}
\paragraph*{Phantom tests}
The accuracy of the US/MR image fusion has been tested on an abdominal phantom CIRS Model 057. The phantom tests have been conducted on the CT (Fig.~\ref{fig:result0}) and MR (Fig.~\ref{fig:result0MRI}) images. The skin surface has been acquired by segmenting the CT acquisition of the phantom. In contrast, the 3D surface obtained by the camera was captured by moving the camera around the phantom to simulate a better possible clinical configuration, where the EM transmitter and camera are placed around the patient bed. The accuracy result is better in the 0 degrees and 180 degrees with respect to the lateral one (90-180). In the worst-case scenario, the accuracy error varies from 4.3mm to 13 mm.
\section{Conclusions and future work\label{sec:FUTURE-WORK}}
This paper presents a method for fusing volumetric anatomical images (MRI/CT) with US images through a 3D depth sensor. The main novelty in the fusion between the two images is the co-registration between the skin surface extracted from the volumetric image and the skin surface acquired by the 3D camera. This co-registration, together with the magnetic tracking system and the 3D sensors placed on the probe and camera, allows the fusion of the MRI/CT image with real-time US acquisitions without using external physical markers. The co-registration has satisfactory accuracy and robustness to noise, virtual landmark misalignment, camera acquisition distance, and camera tilting during the acquisition phase.

%The test on volunteers and in the clinical environment demonstrated that the 3D camera acquisition and co-registration streamline the fusion procedure between the US and anatomical images. This enhancement reduces the steps required for fusion imaging, making it accessible even for less experienced operators.
The precision achieved on tests of the complete system integrated within a US system is of the order of the millimetre. In some cases, the dataset has limitations due to the shape (e.g., symmetry of the phantom, anisotropic room illumination).% Moreover, in some tests, the MR acquisition date and the US examination were taken more than ten years apart, increasing morphological changes.

Future work will focus on reducing the computational time by sub-sampling the volumetric image and through the acquisition of the patient’s skin by a 3D camera with a lower resolution. Skin reference has demonstrated great relevance in many other applications such as on breast tissue, vessels and blood evaluation~\cite{jermyn2013fast, wang2012fully},~\cite{lee2018automated}, or in the neurological field for surgical navigation system optimisation and registration~\cite{AnAut4958341:online, beare2016automated}, as well as on the abdominal district~\cite{weston2019automated, baum2012does}. Thus, future works will focus on improving and developing the skin segmentation that, at the moment, provides promising results in accuracy and generality, as well as its visualisation for clinical applications such as surgical intervention planning. Moreover, we will focus on an augmented system to visualise the error and to represent possible misalignment in the volume image other than on the surface.

A potential avenue for further improvement in the coregistration involves exploring camera registration while partially dressing the patient. Further enhancement could include making the image fusion system independent from the patient's position and breathing phase, presenting opportunities for continued refinement in future iterations.
	
	\paragraph{Acknowledgments}
	This work has been supported by the European Commission, NextGenerationEU, Missione 4 Componente 2, “Dalla ricerca all’impresa”, Innovation Ecosystem RAISE “Robotics and AI for Socio-economic Empowerment”,
	ECS00000035.
	\bibliographystyle{alpha}
	\bibliography{sample}
\end{document}